\documentclass[runningheads]{llncs}

\usepackage{eccv}

\usepackage{eccvabbrv}

\usepackage{graphicx}
\usepackage{booktabs}

\usepackage[accsupp]{axessibility}  

\usepackage{hyperref}

\usepackage{orcidlink}

\usepackage{amsmath}
\usepackage{amsfonts}

\usepackage{subcaption}

\usepackage{pgfplots}
\pgfplotsset{compat=newest}

\usepackage{longtable}
\usepackage{multirow}
\usepackage{changepage}
\usepackage{makecell}
\usepackage{tablefootnote}

\usepackage{csquotes}

\usepackage{listings}
\usepackage{color}
\definecolor{backcolor}{RGB}{250, 250, 250}
\definecolor{cmtcolor}{RGB}{140, 250, 140}
\definecolor{kwcolor}{RGB}{60, 150, 240}
\definecolor{numcolor}{RGB}{120, 120, 120}
\definecolor{strcolor}{RGB}{240, 210, 60}
\usepackage{cite}

\usepackage{comment}
\usepackage{xcolor}
\usepackage{soul}

\newcommand{\conf}[1]{\scriptsize{$\pm$#1}}

\begin{document}

\title{CA3D: Convolutional-Attentional 3D Nets for Efficient Video Activity Recognition on the Edge} 

\titlerunning{Convolutional-Attentional 3D Nets}

\author{Gabriele Lagani\inst{1}\orcidlink{0000-0003-4739-5778} \and
Fabrizio Falchi\inst{1}\orcidlink{0000-0001-6258-5313} \and
Claudio Gennaro\inst{1}\orcidlink{0000-0002-3715-149X} \and
Giuseppe Amato\inst{1}\orcidlink{0000-0003-0171-4315}}

\authorrunning{G.~Lagani et al.}

\institute{CNR-ISTI, Pisa (PI) 56124, IT 
\email{\{gabriele.lagani, fabrizio.falchi, claudio.gennaro, giuseppe.amato\}@isti.cnr.it}}

\maketitle

\begin{abstract}
In this paper, we introduce a deep learning solution for video activity recognition that leverages an innovative combination of convolutional layers with a linear-complexity attention mechanism. 
Moreover, we introduce a novel quantization mechanism to further improve the efficiency of our model during both training and inference.
Our 
model maintains a reduced computational cost, while preserving robust learning and generalization capabilities.
Our approach addresses the issues related to the high computing requirements of current models, with the goal of achieving competitive accuracy on consumer and edge devices, enabling smart home and smart healthcare applications where efficiency and privacy issues are of concern.
We experimentally validate our model on different established and publicly available video activity recognition benchmarks, improving accuracy over alternative models at a competitive computing cost.
\end{abstract}

\section{Introduction}

State-of-the-art deep learning models for video processing have shown impressive results in modeling, interpreting, and performing inference on temporal sequences of video frames \cite{tran2018, xie2018, feichtenhofer2019, feichtenhofer2020, sharir2021, bertasius2021, arnab2021, liu2022a, piergiovanni2023}. However, such models suffer from a number of issues related to computational efficiency, both at inference and training time, as well as energy costs \cite{badar2021}.
In terms of computational efficiency, recent models are becoming more and more complex, making it hard to deploy them on commodity hardware, for consumer-oriented and privacy-oriented applications on the edge.
Similarly, in terms of training efficiency, state-of-the-art models require large amounts of GPU hours (hence energy costs) to achieve good task performances \cite{badar2021}. Moreover, such models are generally pre-trained on large collections of data, and then fine-tuned on task-specific data, representing an additional cost in terms of data efficiency \cite{lee2021}.

In video processing applications for environments such as smart home and smart healthcare, privacy represents a major concern, so that it is preferable to keep personal user data within the local environment. In this scenario, it is desirable to have deep learning models for video processing that are accurate enough for the task at hand, possibly leveraging user-specific data for online fine-tuning, while also maintaining a lightweight memory and computing footprint, so that applications can run effectively on edge devices and commodity hardware.
Current state-of-the-art architectures are not well suited for the scenario described so far. On one hand, traditional convolutional network architectures are relatively efficient to run, and show good generalization performances from little amounts of data, thanks to the translation-invariance that is intrinsic in the convolutional layers \cite{tran2018, xie2018, feichtenhofer2019, feichtenhofer2020}; however, the performance that these models achieve, even when training on large collections of data, is overall suboptimal. On the other hand, more recent attention-based architectures \cite{sharir2021, bertasius2021, arnab2021, liu2022a, piergiovanni2023} stemming from the successful Transformer model \cite{vaswani2017}, are able to achieve overall better performances, thanks to the capability of the attention layer to model global relationships between different parts of the input (such as frames that are far apart in time, while convolutions can only focus on local relationships); however, there are a number of drawbacks: 1) the attention computation is costly, 2) training such models to optimal performance requires huge data availability to achieve good generalization, and 3) the models are too complex for commodity hardware and edge devices.

\begin{figure}[t]
    \centering
    \includegraphics[width=0.7\textwidth]{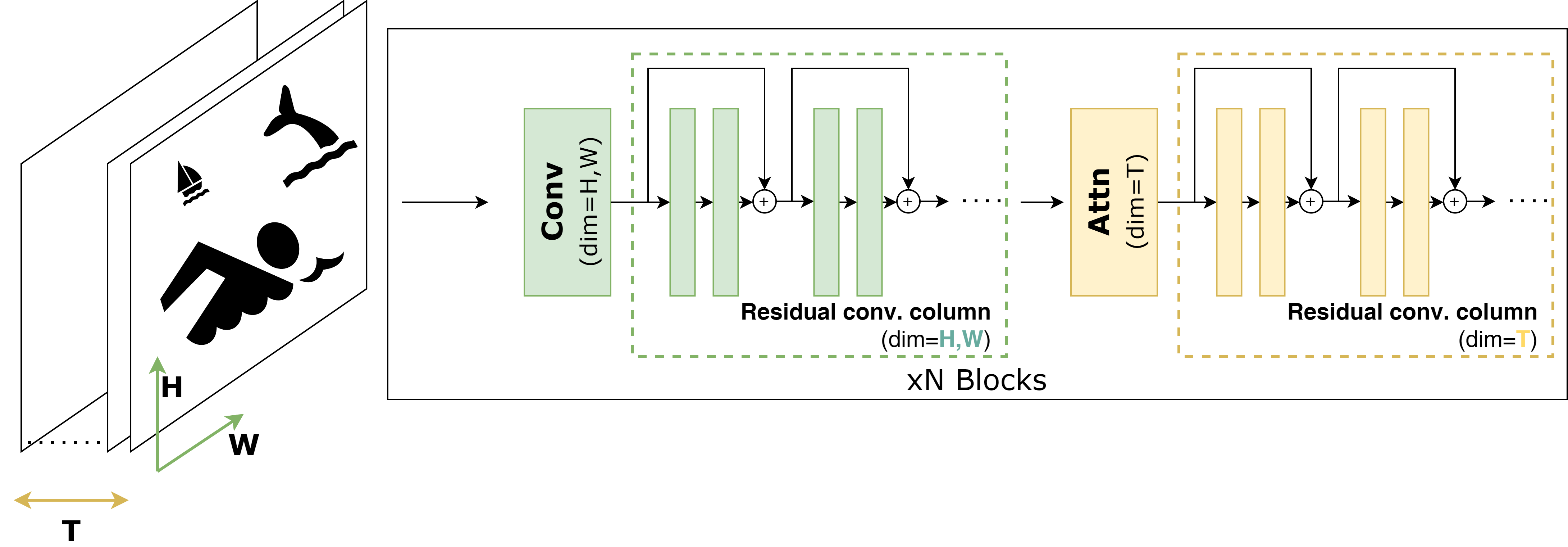}
    \caption{Design of the Convolutional-Attentional 3D (CA3D) neural network for video processing, as a series of Convolutional-Attentional Spatio-Temporal (CAST) blocks. In CAST blocks, convolutional layers are alternated with attention layers, to take advantage of both types of processing. Convolutions are applied along the spatial dimensions, while attention aggregates global information from different frames along the temporal dimension. Processing of each layer is further enhanced with a deep column of residual stages.}
    \label{fig:ca3d}
\end{figure}

In order to overcome these challenges, we designed a novel deep learning block for video processing, named \textit{Convolutional-Attentional Spatio-Temporal} (CAST) layer
. The design of this block is carefully chosen in order to take the best advantages from both types of processing (convolutional and attention), while minimizing the impact on performance and training cost. In particular, we defined a novel linear-complexity attention block, based on local attention windows in the same spirit as \cite{liu2021}, to further reduce the cost of attention computation. We also designed a novel architecture, based on CAST layers, named \textit{Convolutional-Attentional 3D} (CA3D) network (Fig. \ref{fig:ca3d}), together with a quantization mechanism that allows us to reduce the computational burden during both training and inference. 
In order to validate the proposed architecture, we performed experiments on three publicly available and widely used video activity recognition benchmarks, namely UCF101 \cite{ucf}, HBDM51 \cite{hmdb}, and Kinetics400 \cite{kinetics}, comparing our method with several state-of-the-art models and under different quantization settings.
By combining the three proposed ingredients -- 1) combination of convolutional and attentional processing, 2) linear-complexity attention and, 3) our quantization mechanism -- the
overall model is able to run on commodity hardware, for both training and inference, while at the same time achieving performance competitive with state-of-the-art on video activity recognition benchmarks, when considering no additional input streams besides raw RGB frames, and no pre-training on large external datasets. This makes the resulting model suitable for real time video processing applications on the edge.

In summary, our contributions are the following:
\begin{itemize}
    \item We design a novel Convolutional-Attentional 3D (CA3D) neural network model, based on the proposed Convolutional-Attentional Spatio-Temporal (CAST) layers, leveraging both convolutions and attention for efficient video activity recognition, with effective learning and generalization capabilities;
    \item In order to guarantee efficient processing in attention blocks, we define a specific linear-complexity attention module based on local attention windows; 
    \item We introduce a novel quantization mechanism, that allows to further reduce the memory and computing footprint of our model, not only at inference time, but also during training.
\end{itemize}

In the following, we provide a background on existing deep learning solutions for video processing, describing the limitations of current solutions (Section \ref{sec:rel_work}), and we provide the details of our design and  our quantization strategy (Section \ref{sec:method}). We provide an experimental validation of our design (Section \ref{sec:experiments}), and finally we conclude with some remarks and possible future directions (Section \ref{sec:conclusions}).
The code to reproduce our experiments is publicly available. \footnote{\texttt{github.com/GabrieleLagani/ElderlyActivityRecognition}}

\section{Related Work} \label{sec:rel_work}


Early CNN architectures for video data extended the 2D convolutions used for image data to 3D convolutions, in order to process spatio-temporal video data, as in \cite{ji2012, karpathy2014}. One such architecture was the C3D model \cite{tran2015}, using 3D convolutional blocks for video activity recognition. This model was further improved with the introduction of residual blocks, in the R3D architecture \cite{tran2018}. In the same contribution \cite{tran2018}, the authors proposed the R2+1D architecture, factorizing the spatio-temporal 3D convolutional kernel in a 2D spatial convolution, followed by a 1D temporal convolution, in order to achieve a more efficient processing without sacrificing accuracy. 
Other works explored architectural variations for improved performance and efficiency, such as S3D \cite{xie2018}, Multi-Fiber Networks \cite{chen2018a}, Non-Local blocks \cite{wang2018}, or X3D \cite{feichtenhofer2020}. 
At the same time, researchers investigated the possibility to transfer features learned from 2D image data in the context of 3D convolutional networks. The I3D model \cite{kinetics} addresses this idea by \textit{inflating} learned 2D convolutional kernels along the temporal dimension, and then fine-tuning the resulting model on the desired task.

More recent models explored the use of attention mechanisms and transformer-based architectures \cite{vaswani2017}, for spatio-temporal video processing. 
STAM \cite{sharir2021} imediately transfered pre-trained Vision Transformers (ViT) \cite{dosovitskiy2020} to video data by using a ViT pre-trained on images to extract features from each frame in the video, followed by another stack of Transformer layers for feature aggregation across frames. 
Other approaches, such as VidTr \cite{zhang2021} or TimeSFormer \cite{bertasius2021} extended the attention mechanism to the spatio-temporal domain. However, given the complexity of the attention operation, the authors investigated more efficient implementations based on the idea of factorizing the attention block over the spatial and temporal domains. A variety of spatio-temporal attention mechanisms were explored also in ViViT models \cite{arnab2021}, including the factorized attention model, and the factorized encoder model, with a spatial Transformer-based encoder followed by a temporal one (in the same spirit as STAM). 
TubeViT \cite{piergiovanni2023} 
samples 3D patches (or \textit{tubelets}) from the video feature block, at sparse offsets, using various spatial and temporal scales (in such a way that, when the spatial resolution is increased, temporal resolution is decreased, and vice-versa), so that the whole input can be processed at a reduced computational cost.

While Transformer-based architectures leverage the attention mechanism to model global relationships in the spatio-temporal feature maps, they also lose the architectural inductive bias of convolutions, which are typically helpful to improve generalization through translation invariance and local processing. As a result, Transformers are extremely data-hungry, requiring pre-training on very large image or video datasets, in order to learn useful visual features that can then be reused in downstream video tasks, achieving good performances. 
To address the quadratic complexity limitation of the attention operator, more efficient alternatives are receiving attention from researchers, such as sparse attention approximations \cite{child2019}, approximations based on random projections such as the Performer \cite{choromanski2020}, low-rank approximations such as the LinFormer \cite{wang2020}, or attentional processing applied only on local windows as in the Swin Transformer \cite{liu2021}. 
However, even considering optimized variants of the attention mechanisms, Transformer-based approaches still impose extreme memory and computing requirements, both for inference and, in particular, for training. Moreover, the training cost of such models becomes extremely burdening, as large training datasets are required for good performance.
Therefore, the applicability of such models to realistic scenarios, such as smart home or smart healthcare applications on edge devices, remains limited. This is particularly true when local fine-tuning of the system on privacy-constrained user data is necessary, or when real-time constraints are involved. Instead, convolutional models enable efficient processing, but they lack a mechanism for capturing semantic relationships at a global scale. 

A popular approach towards reducing the computational complexity of large models is neural network quantization \cite{gholami2022}. Static post-training quantization \cite{banner2019, choukroun2019} is such a strategy, where training is performed in full precision, e.g. on 32 bits floating point representations, and then the resulting model parameters are converted to a reduced precision, such as 16 bits or less. Dynamic quantization \cite{choi2018, zhang2018} proceeds in a similar spirit, but adopting strategies to adjust quantization ranges adaptively at inference time, based on the distribution of the observed values to be quantized. Typically, however, quantized networks typically suffer from a loss in performance compared to the full-precision models, and further fine-tuning is generally required after quantization. Quantization-Aware Training (QAT) \cite{krishnamoorthi2018, jacob2018} addresses this issue by explicitly targeting the quantized model performance during training: the quantization process of weights and activations is simulated during training, so that the training procedure directly optimizes the performance of the quantized network. In the following, we will focus on QAT as a baseline method for quantization.

Other approaches leverage extra features in addition to RGB frames, i.e. optical flow features extracted from video frames, thus enhancing the video processing architectures with multiple streams of information \cite{simonyan2014, fan2018, sun2018, ng2018, zhu2019, crasto2019, stroud2020, feichtenhofer2019}. However, the additional preprocessing also has a significant performance cost, thus precluding real-time applications. Therefore, it is more desirable to avoid such features, be it for inference, but also (preferably) for training.

Compared to previous approaches, we aim at designing an architecture that effectively mixes convolutional and attentional characteristics, in order to achieve efficient processing, as well as a strong generalization capacity. Moreover, we aim at developing a model that can be effectively trained or fine-tuned without relying on expensive pre-training on extra training data, while also using only raw RGB frames, without requiring costly computations of additional features.
In addition, an efficient attention mechanism based on local attention windows is defined, to avoid the quadratic complexity cost, and a quantization mechanism that allows to perform training directly on 16 bits, without relying on 32 bit representations at any point (while QAT still maintains the true parameters in full precision while learning, with a consistent memory footprint).

\section{Method} \label{sec:method}

In this Section, we present our CA3D network architecture for video activity recognition based on CAST blocks, which combine convolutional and attentional mechanisms, and further enhanced with our linear-complexity attention mechanism, and our novel quantization mechanism, that allows us to reduce the computing footprint of the model, during both inference and training.

\subsection{Efficient Convolutional-Attentional Network for 3D Spatio-Temporal Data}

The proposed CAST block alternates convolutional layers with attention-based processing layers, as shown in Fig. \ref{fig:ca3d}. Convolutions are performed over the spatial dimensions of the feature maps, and their purpose is to extract visual features at a higher level of complexity as depth increases. At the same time, convolutional layers leverage locality and translation invariance as an effective inductive bias for generalization. However, since convolutional layers only process information based on local regions of the input feature map, they are unable to model long-range dependencies, such as those between distant frames of a spatio-temporal feature map. Instead, for this purpose, we introduce attention layers in our design. However, since attention is computationally expensive, we apply the attention mechanism only along the temporal dimension. This is opposed to previous architectures, which leveraged the attention mechanism along both spatial and temporal dimensions, resulting in a high computational demand \cite{sharir2021, bertasius2021, arnab2021, piergiovanni2023}. We experimentally observed that the attention computation can be limited to the temporal dimension, while convolutions are sufficient for the spatial dimension, without significantly affecting task performance, but resulting in a significant improvement in terms of computing efficiency. Moreover, since the attention mechanism grows quadratically in complexity with the sequence length, we designed instead an efficient variant of the full attention mechanism that can lead to comparable results with linear complexity. This variant is inspired by the model used in the Swin Transformer \cite{liu2021, liu2022a}, where attention is applied only within local windows extracted from the input feature map, following a grid pattern, and successive attention layers shift the windows by a small offset, so that information can be propagated across different windows as well. 

The attention strategy that we propose still considers local windows, but each window is centered around an input token. Hence, we consider a window for each possible offset in the feature map, instead of following a fixed grid. Each token only attends to the tokens in its neighborhood, as defined by the local window. In this case, global information propagation is still achieved over successive layers, because the attention output at a given token position of a given layer will depend on all the tokens in the neighborhood, which in turn depends on their respective neighborhoods in the previous layers, thus enlarging the receptive field with each layer. 
The remaining part of the attention block still corresponds to a standard Multi-Head Self-Attention (MHSA) \cite{vaswani2017}, with Query-Key-Value (QKV) mappings achieved through linear layers, and additive positional embeddings applied on the input tokens.

Both
the convolutional processing layer and the attention layer in CAST blocks are followed by a deep column of residually connected convolutional layers \cite{he2016a} (convolutions are performed over height and width dimensions for the spatial processing part, and over the temporal dimension for the temporal processing part). These correspond to standard residual blocks for convolutional processing, while for the attentional processing part, these columns essentially replace the role of Multi-Layer Perceptrons (MLP) for feature mixing used in transformers \cite{vaswani2017, dosovitskiy2020}, which are indeed equivalent to unitary-width convolutions. One of the advantages of this scheme is that we can extend token mixing also to broader convolutions, thus further enhancing information propagation along a desired dimension. Moreover, we always represent tensors as spatio-temporal token maps, so that we never lose the neighborhood relationships among tokens in space or time, giving a powerful inductive bias for generalization. Specifically, we used columns with 2 residual stages, corresponding to 4 convolutional layers each.
Every convolutional layer is followed by a ReLU nonlinearity \cite{nair2010}, and the output is normalized by Batch Normalization (BatchNorm) \cite{ioffe2015}. The latter is also the type of normalization that we adopted in our attention blocks, as this configuration appeared to be more stable during training, compared to other alternatives (such as LayerNorm \cite{ba2016}). Residual blocks follow the optimized structure proposed in \cite{he2016b}.
The CA3D network is composed by a series of CAST blocks. We do not use a class token in our design, as the necessary information for classification is already contained in the other tokens. Instead, a final classifier, composed by a global average pooling \cite{gholamalinezhad2020} and a linear layer, is applied after the sequence of CAST blocks, finally obtaining the class scores. Dropout \cite{srivastava2014} with rate 0.5 is applied before the final layer.

\subsubsection{Details of the CA3D architecture.}

\begin{figure}
    \centering
    \includegraphics[width=\textwidth]{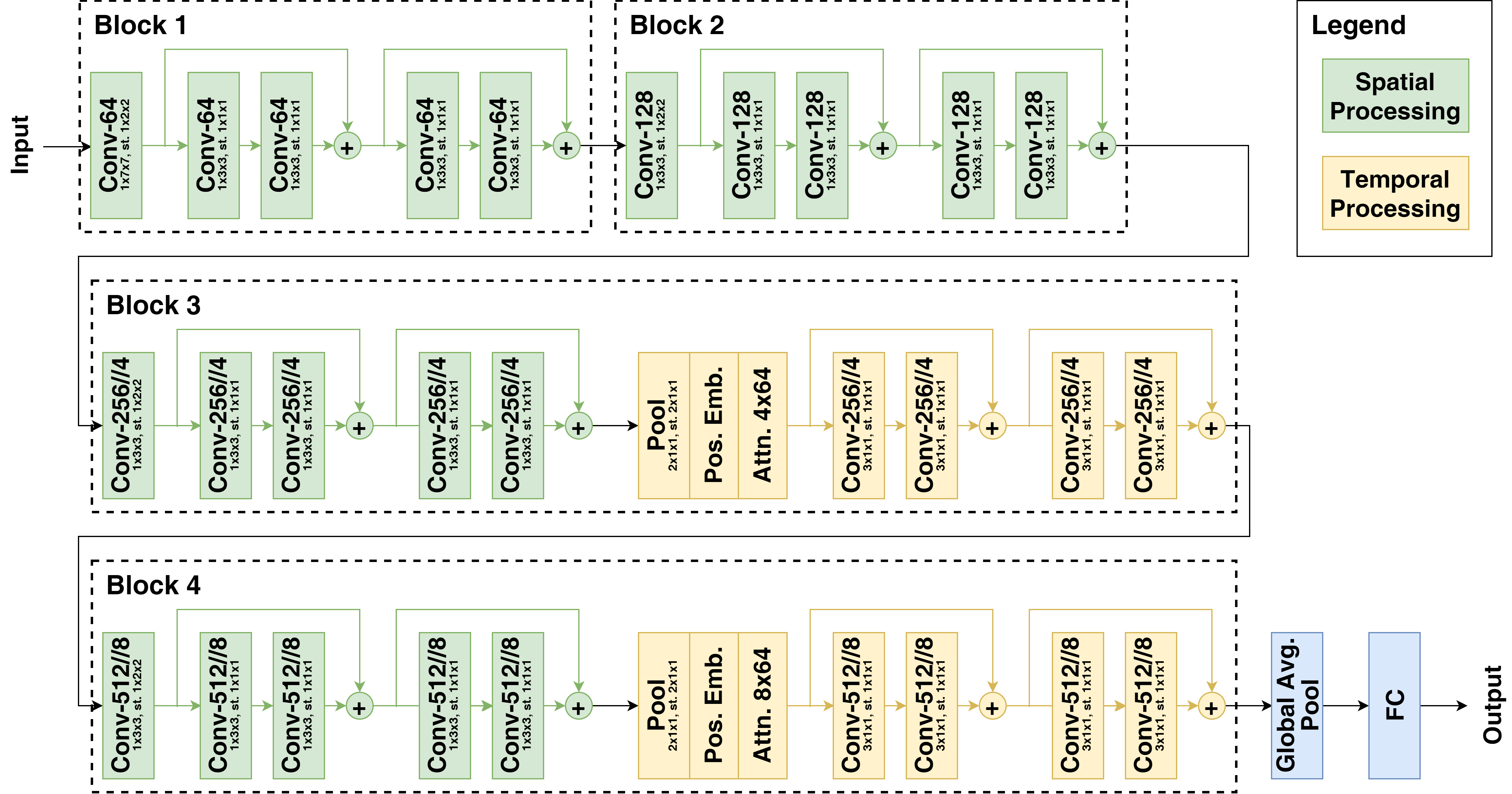}
    \caption{Structure of the CA3D architecture in terms of CAST blocks. Blocks are formed by spatial and temporal processing parts. The spatial part is implemented in terms of convolutional blocks, followed by a column of residually connected layers. The temporal part is mediated by an attention operator, which is followed again by a column of residual layers. Pooling layers shrink the size of the tensor along the temporal dimension. }
    \label{fig:ca3d_blocks}
\end{figure}

The specific CA3D model that we propose, also
depicted in Fig. \ref{fig:ca3d_blocks},
is structured as follows. 
The first CAST block uses a 7x7 spatial convolution with stride 2, followed by a column with 2 residual stages, for a total depth of 5 layers, and 64 channels. 
The temporal part for the first block has depth 0, i.e. it is empty; this is because earlier layers tend to focus on purely spatial features, while abstract temporal features only emerge at deeper layers \cite{sharir2021, arnab2021}. 
The second CAST block has a similar structure, but the spatial convolution has size 3x3 and stride 2, and there are 128 channels. 
The third block uses a spatial convolution of size 3x3 and stride 2, with 256 channels, followed by a column with 2 residual stages; for the temporal part, the attention block divides the 256 channels in 4 heads of 64 channels each, followed again by a 2 residual stage column. This amounts to a total of 10 layers (counting attention as one layer, while ignoring nonlinearities and normalization layers from the count). 
The fourth block has a similar structure, but it uses 512 channels, divided in 8 heads of 64 channels each, and a 2 residual stage column for the spatial and the temporal part, for a total of 10 more layers. 
Residual blocks all use padded convolutions of size 3 and stride 1. 
Furthermore, the temporal dimension is compressed after the second and third block by a max pooling of size 2 and stride 2. Together with a final linear classifier, this amounts to a 31 layers architecture, with $\sim$7M parameters, which is suitable for running on commodity hardware both during inference, and also (thanks to our quantization strategy discussed hereafter) during training. 
In the Suppl. Material, the reader can find more information about hyperparameter configuration
, as well as additional experimental results focusing on an ablation study, to clarify the impact of different architectural components on the CA3D architecture
.

\subsection{A Quantization Mechanism for Efficient Training and Inference}

Efficient training and inference are achieved by further introducing a novel quantization mechanism that maps floating point number representations from 32 bits (\texttt{float32}) to 16 bits (\texttt{float16}), leading to an improvement in memory footprint by a factor of 2, and to faster processing as well. 
We could use static \cite{banner2019, choukroun2019, gholami2022} or dynamic \cite{choi2018, zhang2018} quantization to convert model parameters and feature maps from \texttt{float32} to \texttt{float16}. These approaches is effective on pre-trained models, followed by some extra fine-tuning in the new \texttt{float16} regime, in order to improve performance at inference time. However, the pre-training must be performed in \texttt{float32}, otherwise numerical instabilities arise. A more sophisticated scheme is Quantization-Aware Training (QAT) \cite{krishnamoorthi2018, jacob2018}. In this case, the parameters are maintained on 32 bits during training. However, for each training pass, conversion to 16 bits is simulated during training, in order to reproduce the effect of quantization errors. This allows to find trained models that, once quantized, are able to achieve near-lossless performance compared to the non-quantized models. Indeed, by maintaining the parameters on 32 bits, numerical stability of gradient computation and optimization can still be maintained during training; however, while this approach improves the model footprint at inference time, training costs are still high. 

We designed a quantization mechanism that can improve the models both during training and inference, where both phases are implemented in \texttt{float16}, and at no point we rely on \texttt{float32} representations. The idea is that, instead of performing optimization in a space of parameters $\mathbf{w}$ (represented in \texttt{float16}) where optimization is unstable, we perform optimization in another space of pre-parameters $\mathbf{\theta}$ where optimization is easier. This is done by defining a mapping $\mathbf{f}: \mathbf{\theta} \rightarrow \mathbf{w}$ so that the weights $\mathbf{w}$ of our network are generated from the pre-parameters $\mathbf{\theta}$ according to mapping $\mathbf{f}$. By appropriately choosing mapping $\mathbf{f}$, optimization can be made more stable in the space of pre-parameters $\mathbf{\theta}$. 
We argue that this methodology could be used, in the future, in order to address other optimization problems where a problem-specific mapping can cast the loss landscape to a more convenient configuration.

In our case, instabilities can occur because large values for parameters and gradients, close to the upper bound of the representation range (i.e. 32768), may arise. Hence, we define a mapping 
\begin{equation} \label{eq:preparam}
    \mathbf{w} = \mathbf{f}(\mathbf{\theta}) = \mathbf{\theta} / T,
\end{equation} 
where $T$ is a constant. If, for example, we set $T = 0.1$, we are able to implicitly shift the representation range from $[1/32768, 32768]$ to $[1/3276.8, 327680]$. The latter range is much more useful to represent the actual numbers that appear in our problem, 
making optimization easier in this space.
Finally, in the backward phase, parameter gradients ($\texttt{grad}_{\mathbf{w}}$) are transferred to the gradient buffers of the pre-parameters ($\texttt{grad}_{\mathbf{\theta}}$) by a backward mapping aimed at conserving the relative magnitude of gradients w.r.t. the optimization variables:
\begin{equation} \label{eq:preparam_bw}
    \texttt{grad}_{\mathbf{\theta}} = T \cdot \texttt{grad}_{\mathbf{w}} .
\end{equation}
After that, the optimizer step is invoked on the pre-parameters, and their new values are used to compute the updated parameters through Eq. \ref{eq:preparam}.

\section{Experiments} \label{sec:experiments}

We performed experiments on three public and widely used video activity recognition benchmarks: 
UCF101 \cite{ucf}, HMDB51 \cite{hmdb}, and Kinetics400 \cite{kinetics}.

Our experiments are structured in three stages: first, we show a comparison of CA3D and other state-of-the-art architectures on UCF101 and HMDB51, while also comparing different types of quantization; second, we compare our approach with other methods on the Kinetics400 dataset; third, we show a comparison of different approaches in terms of computational footprint. It is important to point out that, in our experiments, we always considered training conditions in which 1) only raw RGB frames are used, with \textit{no additional preprocessing} (besides standard data augmentation), and 2) \textit{no additional external training data} are used. This conditions were kept the same across all the experiments, in order to make comparisons on equal footings. This is different from previous approaches relying on additional feature extraction/additional streams in the architecture \cite{fan2018, sun2018, ng2018, zhu2019, crasto2019, stroud2020}, and approaches relying on pre-training on large image datasets \cite{kinetics, feichtenhofer2020, xie2018, sharir2021, bertasius2021, arnab2021, liu2022a, li2022a, li2022b, piergiovanni2023}.
In the following, we report the test split results of our experimental scenarios in the various tasks. Details about the experimental conditions are reported in the Suppl. Material, as well as additional hyperparameter search and ablation results highlighting the impact of different architectural choices on the proposed model's performance.

\subsubsection{UCF101 and HMDB51.}

\begin{table}[!tb]
  \caption{Test accuracy of different methods under different types of quantization on UCF101 and HMDB51. \underline{Underlined} results represent the best type of quantization for each architecture, while \textbf{bold} represents the best result. It can be observed that our CA3D architecture performs comparably or even better than previous methods in our scenarios, effectively combining the advantages of convolutional and attentional processing. Moreover, our quantization mechanism achieves comparable results w.r.t. the unquantized models and QAT in almost all the cases. In some scenarios, quantization errors are even helpful in inducing better generalization, often improving over the unquantized baselines.}
  \label{tab:acc}
  \begin{subtable}[t]{0.5\textwidth}
  \caption{Test accuracy on UCF101.}
  \label{tab:acc_ucf}
  \scriptsize
  \begin{tabular}{@{}c|c|c@{}}
    \toprule
    \textbf{Model} & \textbf{Quantization} & \textbf{Acc. (\%)} $\uparrow$ \\
    \midrule
    \multirow{3}{*}{R3D-R18 \cite{tran2018}} 
        & \texttt{float32} & 90.3 \conf{1.4} \\ 
        & \texttt{QAT} & \underline{91.7} \conf{1.3} \\ 
        & \textbf{Ours} (\texttt{float16}) & 89.7 \conf{1.4} \\ 
    \hline
    \multirow{3}{*}{R2+1D-R18 \cite{tran2018}} 
        & \texttt{float32} & 85.6 \conf{1.6} \\ 
        & \texttt{QAT} & \underline{94.1} \conf{1.1} \\ 
        & \textbf{Ours} (\texttt{float16}) & 93.2 \conf{1.2} \\ 
    \hline
    \multirow{3}{*}{I3D \cite{kinetics}} 
        & \texttt{float32} & 90.8 \conf {1.3} \\ 
        & \texttt{QAT} & 91.5 \conf{1.3} \\ 
        & \textbf{Ours} (\texttt{float16}) & \underline{92.2} \conf{1.2} \\ 
    \hline
    \multirow{3}{*}{X3D-XL \cite{feichtenhofer2020}} 
        & \texttt{float32} & 77.0 \conf{1.9} \\ 
        & \texttt{QAT} & 80.1 \conf{1.8} \\ 
        & \textbf{Ours} (\texttt{float16}) & \underline{86.4} \conf{1.6} \\ 
    \hline
    \multirow{3}{*}{STAM-B \cite{sharir2021}} 
        & \texttt{float32} & 81.2 \conf{1.8} \\ 
        & \texttt{QAT} & \underline{86.5} \conf{1.6} \\ 
        & \textbf{Ours} (\texttt{float16}) & 81.0 \conf{1.8} \\ 
    \hline
    \multirow{3}{*}{TimesFormer-B \cite{bertasius2021}} 
        & \texttt{float32} & \underline{84.1} \conf{1.6} \\ 
        & \texttt{QAT} & 80.5 \conf{1.8} \\ 
        & \textbf{Ours} (\texttt{float16}) & 80.3 \conf{1.8} \\ 
    \hline
    \multirow{3}{*}{ViViT-2 \cite{arnab2021}} 
        & \texttt{float32} & 77.1 \conf{1.9} \\ 
        & \texttt{QAT} & \underline{85.5} \conf{1.6} \\ 
        & \textbf{Ours} (\texttt{float16}) & 76.9 \conf{1.9} \\ 
    \hline
    \multirow{3}{*}{Swin3D-T \cite{liu2022a}} 
        & \texttt{float32} & 79.3 \conf{1.9} \\ 
        & \texttt{QAT} & \underline{83.5} \conf{1.7} \\ 
        & \textbf{Ours} (\texttt{float16}) & 80.5 \conf{1.8} \\ 
    \hline
    \multirow{3}{*}{TubeViT-B \cite{piergiovanni2023}} 
        & \texttt{float32} & 80.9 \conf{1.8} \\ 
        & \texttt{QAT} & \underline{87.2} \conf{1.5} \\ 
        & \textbf{Ours} (\texttt{float16}) & 80.9 \conf{1.8} \\ 
    \hline
    \multirow{3}{*}{\textbf{CA3D (Ours)}} 
        & \texttt{float32} & 94.1 \conf{1.1} \\ 
        & \texttt{QAT} & \textbf{\underline{94.8}} \conf{1.1} \\ 
        & \textbf{Ours} (\texttt{float16}) & \textbf{\underline{94.8}} \conf{1.1} \\ 
  \bottomrule
  \end{tabular}
  \end{subtable}
  ~
  \begin{subtable}[t]{0.5\textwidth}
  \caption{Test accuracy on HMDB51.}
  \label{tab:acc_hmdb}
  \scriptsize
  \begin{tabular}{@{}c|c|c@{}}
    \toprule
    \textbf{Model} & \textbf{Quantization} & \textbf{Acc. (\%)} $\uparrow$ \\
    \midrule
    \multirow{3}{*}{R3D-R18 \cite{tran2018}} 
        & \texttt{float32} & 51.4 \conf{3.3} \\ 
        & \texttt{QAT} & \underline{55.4} \conf{3.2} \\ 
        & \textbf{Ours} (\texttt{float16}) & 52.3 \conf{3.3} \\ 
    \hline
    \multirow{3}{*}{R2+1D-R18 \cite{tran2018}} 
        & \texttt{float32} & 46.6 \conf{3.3} \\ 
        & \texttt{QAT} & \underline{60.7} \conf{3.2} \\ 
        & \textbf{Ours} (\texttt{float16}) & 57.0 \conf{3.3} \\ 
    \hline
    \multirow{3}{*}{I3D \cite{kinetics}} 
        & \texttt{float32} & 58.3 \conf{3.2} \\ 
        & \texttt{QAT} & \underline{58.8} \conf{3.2} \\ 
        & \textbf{Ours} (\texttt{float16}) & 57.3 \conf{3.2} \\ 
    \hline
    \multirow{3}{*}{X3D-XL \cite{feichtenhofer2020}} 
        & \texttt{float32} & 30.5 \conf{3.0} \\ 
        & \texttt{QAT} & 35.1 \conf{3.1} \\ 
        & \textbf{Ours} (\texttt{float16}) & \underline{37.3} \conf{3.2} \\ 
    \hline
    \multirow{3}{*}{STAM-B \cite{sharir2021}} 
        & \texttt{float32} & 45.6 \conf{3.3} \\ 
        & \texttt{QAT} & \underline{48.7} \conf{3.3} \\ 
        & \textbf{Ours} (\texttt{float16}) & 44.5 \conf{3.2} \\ 
    \hline
    \multirow{3}{*}{TimesFormer-B \cite{bertasius2021}} 
        & \texttt{float32} & 35.4 \conf{3.1} \\ 
        & \texttt{QAT} & \underline{48.7} \conf{3.3} \\ 
        & \textbf{Ours} (\texttt{float16}) & 40.2 \conf{3.2} \\ 
    \hline
    \multirow{3}{*}{ViViT-2 \cite{arnab2021}} 
        & \texttt{float32} & 27.1 \conf{2.9} \\ 
        & \texttt{QAT} & \underline{41.2} \conf{3.2} \\ 
        & \textbf{Ours} (\texttt{float16}) & 28.3 \conf{2.9} \\ 
    \hline
    \multirow{3}{*}{Swin3D-T \cite{liu2022a}} 
        & \texttt{float32} & \underline{34.5} \conf{3.1} \\ 
        & \texttt{QAT} & 28.7 \conf{3.0} \\ 
        & \textbf{Ours} (\texttt{float16}) & 33.2 \conf{3.1} \\ 
    \hline
    \multirow{3}{*}{TubeViT-B \cite{piergiovanni2023}} 
        & \texttt{float32} & 35.6 \conf{3.1} \\ 
        & \texttt{QAT} & \underline{46.5} \conf{3.3} \\ 
        & \textbf{Ours} (\texttt{float16}) & 36.3 \conf{3.1}  \\ 
    \hline
    \multirow{3}{*}{\textbf{CA3D (Ours)}} 
        & \texttt{float32} & 60.2 \conf{3.2} \\ 
        & \texttt{QAT} &  62.2 \conf{3.2} \\ 
        & \textbf{Ours} (\texttt{float16}) & \textbf{\underline{63.2}} \conf{3.2} \\ 
  \bottomrule
  \end{tabular}
  \end{subtable} 
\end{table}

The first set of experiments that we propose aims at comparing the accuracy of CA3D with other methods on UCF101 (Table \ref{tab:acc_ucf}) and HMDB51 (Table \ref{tab:acc_hmdb}). Moreover, they aim at assessing the impact of different types of quantization for each considered method. First of all, we must point out that the datasets that we have chosen for this first comparison are relatively small, hence they are useful to assess the generalization properties of the models from few training samples. We compare CA3D with CNN-based (R3D with ResNet18 backbone -- R3D-R18 \cite{tran2018}, R2+1D 
with ResNet18 backbone -- R2+1-R18 \cite{tran2018}, I3D \cite{kinetics}, X3D-XL \cite{feichtenhofer2020}) and Transformer-based methods (STAM-B \cite{sharir2021}, TimesFormer-B \cite{bertasius2021}, ViViT model 2 -- ViViT-2 \cite{arnab2021}, Swin3D-T \cite{liu2022a}, TubeViT-B \cite{piergiovanni2023}
). From the results, we can notice that CNN-based methods generalize better that Transformer-based methods in this context. In fact, the generalization difficulties of Transformers with small datasets are well known \cite{lee2021}. Instead, state-of-the-art methods -- both convolutional and attentional -- are typically pre-trained on large image datasets (e.g. \cite{kinetics} or \cite{dosovitskiy2020}), in order to develop preliminary visual features and achieve good generalization results on the downstream video tasks.
Notably, thanks to the convolutional structure of our architecture, our CA3D model maintains the generalization flexibility of CNN-based models; at the same time, the integration of a temporal attention mechanism introduces the capacity to model long-range temporal structure in the data, increasing the overall performance compared to other methods. In summary, the CA3D design is able to effectively take advantage of attention-based processing, improving over purely convolutional baselines, without suffering in terms of generalization as other attention-based models.

Considering different quantization mechanisms, we can observe that training in a quantized (\texttt{float16}) or fake-quantized (\texttt{QAT}) setting often helps improving generalization performance. Comparing our \texttt{float16} quantization approach with the \texttt{QAT} and \texttt{float32} baselines shows that our method generally achieves comparable results, while \textit{never relying on representations other than} \texttt{float16}. Instead, we must point out that raw \texttt{float16} training, without using our methodology, always resulted in \textit{numerically unstable} behavior, with parameters not converging to an optimum, and leading either to infinities or to random-chance results. It should also be noted that, in order to make comparisons on equal footings, the hyperparameters of each experiment were optimized for the \texttt{float32} quantization regime, and then \textit{the same training conditions were maintained} during the \texttt{float16} experiments. This shows the versatility of our quantization mechanism, which does not require specific hyperparameter tweaking. 

\subsubsection{Kinetics400.}

\begin{table}[!tb]
  \caption{Training and test accuracy of different methods on the Kinetics400 dataset. The best result is highlighted in \textbf{bold}. It can be observed that, in our scenario where no additional preprocessing nor pretraining on external data is used, our CA3D architecture performs better than previous methods at test time. Instead, while CNN-based models exhibit competitive generalization capabilities in this task, Transformer-based models are unable to generalize well without previous pre-training. Nevertheless, CA3D effectively combines the advantages of convolutional and attentional processing, together with the proposed optimized structure and quantization mechanism for lightweight processing.}
  \label{tab:kinetics}
  \centering
  \begin{tabular}{@{}c|c|c|@{}}
    \toprule
    \textbf{Model} & \textbf{Train Acc. (\%)} $\uparrow$ & \textbf{Test Acc. (\%)} $\uparrow$ \\
    \midrule
    R3D-R18 \cite{tran2018} & 57.2 \conf{0.7} & 47.5 \conf{0.7} \\
    R2+1D-R18 \cite{tran2018} & 56.5 \conf{0.7} & 49.8 \conf{0.7} \\
    X3D-XL \cite{kinetics} & 60.8 \conf{0.6} & 51.7 \conf{0.7} \\
    STAM-B \cite{sharir2021} & 51.8 \conf{0.7} & 29.9 \conf{0.6} \\
    TimesFormer-B \cite{bertasius2021} & 35.6 \conf{0.6}  & 27.7 \conf{0.6}  \\
    TubeViT-B \cite{piergiovanni2023} & \textbf{67.4 }\conf{0.6} & 42.2 \conf{0.7} \\
    \hline
    \textbf{CA3D (Ours)} & 58.2 \conf{0.6} & 49.9  \conf{0.7} \\ 
    \textbf{CA3D-L (Ours)} & 64.0 \conf{0.6} & \textbf{52.1}  \conf{0.7} \\ 
  \bottomrule
  \end{tabular}  
\end{table}

We performed additional experiments comparing some of the methods as in the previous subsection with the proposed CA3D architecture, combined with our quantization, on a more complex benchmark, i.e. Kinetics400. 
Since this is a more complex task, we also trained a larger CA3D model, namely CA3D-L. This is obtained by increasing the depth of residual columns. Specifically, CA3D-L has a residual column of depth 4, instead of 2, at the third CAST block, and depth 8 at the fourth. 
For our comparison, we considered some CNN-based models, namely R3D-R18, R2+1D-R18, 
X3D-XL and Transformer-based models, namely STAM-B, TimesFormer-B, TubeViT-B. 
Since experiments on Kinetics400 are costly, we chose to focus on a subset of methods which yielded the most promising results in the previous experimental scenarios. At the same time, the compared methods represent a relevant pool of approaches which address the efficiency problem from different perspectives, such as effective tokenization (TubeViT), factorization of spatial-temporal processing (R2+1D, TimesFormer, STAM), or architectural design (R3D, X3D).
Results are reported in Table \ref{tab:kinetics}. We can observe that, also in this case, Transformer-based models find it harder to generalize without prior pre-training on large image datasets. Indeed, the highest training performance is observed with TubeViT, but the lack of generalization is shown by the sub-competitive test result. Generalization appears to be relatively easier for convolutional models. These results confirm, again, that the novel combination of convolutional and attentional processing in CA3D helps to effectively take advantage of attention-based processing, improving over purely convolutional baselines, while maintaining the generalization properties that other attention-based models are lacking. Moreover, CA3D has performance competitive with bigger model, but with a significant performance advantage, as shown in the next subsection.
It can be noticed that other studies report stronger test results on this dataset \cite{piergiovanni2023, bertasius2021, sharir2021, kinetics, tran2018}, but this is motivated by certain differences in the evaluation scenario, e.g. multiple crops used for testing, different input sizes and frame-rates, or pre-training on large non-publicly available datasets. 
Most importantly, test results in literature show comparable results to ours in validation, but higher scores in testing. This is due to the \textit{10-LeftCenterRight} cropping strategy \cite{kinetics} employed at test time, which helps to greatly improve the results. However, this strategy requires 30 times more compute, and it is therefore unfeasible for the targeted scenario of constrained/edge applications. 
Although certain methods can be beneficial for the final results, they can make certain strengths and weaknesses of the compared approaches less evident or unclear, so we prefer to reduce such factors of variation whenever possible. Moreover, since different works use different methodologies, it is not possible to draw comparisons on equal footings. In order to address this issue and make consistent comparisons across different methods, we adopted a common protocol in all scenarios. In particular, we used no additional pre-training, and we only reported single-crop testing results.

\subsubsection{Computing Footprint.}

\begin{table}[!tb]
  \caption{Computing footprint of different models. We report the number of model parameters, the forward pass GFLOPs per single videoclip and per frame, the training memory footprint, and the training processing throughput in frames per second, considering mini-batch processing with 20 video clips of 16 frames. Best results are highlighted in \textbf{bold}. It can be observed that the proposed CA3D architecture, thanks to its optimized structure and combined with our quantization mechanism, is lightweight enough to be suitable for consumer or edge devices, both for inference and, if needed, further training. At the same time, as observed above, the proposed methodology enables effective exploitation of attention-based processing and generalization.}
  \label{tab:footprint}
  \centering
  \begin{tabular}{@{}c|c|c|c|c@{}}
    \toprule
    \textbf{Model} & \textbf{\# Parameters} $\downarrow$ & \textbf{GFLOPs} $\downarrow$ & \textbf{Memory (GB)} $\downarrow$ & \textbf{Frames/s} $\uparrow$ \\
    \midrule
    R3D-R18 \cite{tran2018}  & 33M & 9.9 & 7.4 & 360 \\ 
    R2+1D-R18 \tablefootnote{See Suppl. Material for details and motivations about differences in number of parameters between R2+1D and R3D} \cite{tran2018}  & 15M & 7.4 & 11.6 & 167 \\ 
    X3D-XL \cite{feichtenhofer2020}  & 11M & 11.9 & 12.0 & 402 \\ 
    STAM-B \cite{sharir2021}  & 119M & 21.7 & 11.8 & 182 \\ 
    TimesFormer-B \cite{bertasius2021}  & 121M & 16.9 & 10.6 & 199 \\ 
    TubeViT-B \cite{piergiovanni2023}  & 86M & 9.2 & 5.0 & 417 \\ 
    \hline
    \textbf{CA3D (Ours)}  & \textbf{7M} & \textbf{6.3} & \textbf{4.6} & \textbf{500} \\ 
    \textbf{CA3D-L (Ours)}  & 19M & 9.6 & 6.1 & 280 \\ 
  \bottomrule
  \end{tabular}
\end{table}

In the following, we discuss the compute footprint of different models, comparing performance indicators measured on both convolutional and attention-based architectures 
, reported in Tab. \ref{tab:footprint}
. 
We indicate the number of parameters of each model, the GFLOPs required to process a single video clip of 16 frames and size 112x112 pixels, estimated using the \texttt{perf} Linux utility, the memory footprint during a training pass with mini-batches of 20 samples, and the measured throughput during training, in terms of processed frames per second, on application-oriented consumer hardware (see Suppl. Material).

It can be observed that our model is able to effectively leverage attentional processing, showing good generalization without requiring prior pre-training on large datasets. However, compared to other attention-based models, our design, combined with the proposed quantization mechanism, is more lightweight and suitable for constrained hardware. Compute footprint indicators are comparable to those of some purely convolutional architectures, although the possibility to leverage attention-based computation allows our architecture to improve its modeling capabilities compared to purely convolutional ones, as observed in previous experiments. 

\section{Conclusions, Limitations, Future Work} \label{sec:conclusions}

We presented the Convolutional-Attentional 3D (CA3D) network, a video processing model built using the proposed Convolutional-Attentional Spatio-Temporal (CAST) blocks. Our design is based on three key ingredients: 1) an innovative combination of structured convolutional processing, for better generalization, with attention-based processing, for better performance, 2) a linear-complexity attention mechanism for more efficient processing, and 3) a novel quantization mechanism, to make our architecture more lightweight and suitable for low-end hardware, targeting smart home/smart healthcare applications on the edge. Experimental comparisons on UCF101, HMDB51, and Kinetics400 showed that the proposed model can effectively leverage attention-based processing, improving over purely convolutional baselines, while at the same time achieving better generalization compared to purely attentional methods. Moreover, our overall design, together with the proposed quantization mechanism, allows to reduce the computing footprint of our model. In particular, the proposed quantization leverages a mapping of the model parameters to a different space, where optimization is more stable, so that our model can run, under constrained application scenarios, both during training and inference.

A possible limitation of our work is that, while the compute requirements of CA3D are favourable compared to Transformer-based models, and task performance is higher compared to CNN-based models, the computing footprint w.r.t. some of the convolutional architectures still needs to be improved. This suggest that further computational improvements can be achieved by focusing on the architectural patterns of CNN-based models that have the most influence on the compute profile. Investigation and integration of such patterns in CAST blocks can be a possible future direction. 
Moreover, our quantization mechanism can be further improved by leveraging more sophisticated pre-parameter mappings, and their applicability to other contexts can be explored. 
Finally, implementations of our model in biologically-inspired \textit{neuromorphic} devices \cite{survey_snn, survey_plasticity} can further enhance the applicability of the proposed approach in the real-world. 

\section*{Acknowledgements}
\noindent This work was partially supported by the following projects: \\
- Social and hUman ceNtered XR (SUN) funded by the Horizon Europe Research \& Innovation Programme (Grant N. 101092612), \\
- Tuscany Health Ecosystem (THE) funded by the National Recovery and Resilience Plan (NRPP), within the NextGeneration Europe (NGEU) Programme (CUP I53C22000780001), \\
- AICult project (code 2022M95RC7), funded by the Italian Ministry for University and Research (MUR), under the PRIN 2022 program, \\
- AI4Media, funded by the EC, within the Horizon 2020 Programme (Grant N. 951911), \\
- INAROS (INtelligenza ARtificiale per il mOnitoraggio e Supporto agli anziani) co-funded by Tuscany Region POR FSE (CUP B53D21008060008).

%
%
\bibliographystyle{splncs04}
\bibliography{references}
\end{document}